\documentclass[letterpaper, 10 pt, conference]{ieeeconf}  % Comment this line out if you need a4paper

\IEEEoverridecommandlockouts                              % This command is only needed if 
\usepackage{graphicx} % for pdf, bitmapped graphics files
\usepackage{mathptmx} % assumes new font selection scheme installed
\usepackage{mathtools,mathptmx}
\usepackage{times} % assumes new font selection scheme installed
\usepackage{amsmath,amssymb,amsfonts}
\usepackage{cite}
\usepackage{pgfplots}
\pgfplotsset{compat=1.7}
\usepackage{pgfplotstable}
\usepackage{tikz,pgfplots}
\usepackage{physics} 
\usepackage{siunitx} 
\usepackage{enumerate} 
\usepackage{tikz,pgfplots}
\usepackage{verbatim}
\usepackage{color}
\usepackage{mathtools}
\usepackage{breqn}
\usepackage{subcaption}
\usepackage{array}
\usepackage{makecell}
\usepackage{soul}  % we can delete this package later
\usepackage{comment}
\usepackage{textcomp}
\usepackage{subcaption}
\usepackage{algorithm}
\usepackage{float}
\usepackage{algpseudocode} 
\usepackage{lipsum}
\usepackage{overpic}
\usepackage[labelsep=period]{caption}
\usepackage[colorlinks,citecolor=blue,urlcolor=blue,linkcolor=blue,bookmarks=false,hypertexnames=true]{hyperref}

\newcommand{\rev}[1]{\textcolor{black}{#1}}

\usepackage[]{graphicx}
\title{\LARGE \bf
CHILD (Controller for Humanoid Imitation and Live Demonstration): a Whole-Body Humanoid Teleoperation System %(To be submitted to Humanoids 2025)
}
\author{Noboru Myers$^{*}$, Obin Kwon$^{*}$, Sankalp Yamsani, and Joohyung Kim
\thanks{
$^{*}$These authors equally contributed to this work.
All authors are with the KIMLAB (Kinetic Intelligent Machine LAB), University of Illinois Urbana-Champaign, IL 61801, USA.  Author contact information is {\tt\small \{noborum2, obinkwon, yamsani2, joohyung\}@illinois.edu}}
}

\begin{document}

\maketitle
\thispagestyle{empty}
\pagestyle{empty}

%%%%%%%%%%%%%%%%%%%%%%%%%%%%%%%%%%%%%%%%%%%%%%%%%%%%%%%%%%%%%%%%%%%%%%%%%%%%%%%%
\begin{abstract}
% Teleoperation has been a powerful tool to train policies for manipulation and loco-manipulation tasks. 
Recent advances in teleoperation have demonstrated robots performing complex manipulation tasks. However, existing works rarely support whole-body joint-level teleoperation for humanoid robots, limiting the diversity of tasks that can be accomplished. This work presents Controller for Humanoid Imitation and Live Demonstration (CHILD), a compact reconfigurable teleoperation system that enables joint level control over humanoid robots. CHILD fits within a standard baby carrier, allowing the operator control over all four limbs, and supports both direct joint mapping for full-body control and loco-manipulation. Adaptive force feedback is incorporated to enhance operator experience and prevent unsafe joint movements.
We validate the capabilities of this system by conducting loco-manipulation and full-body control \rev{demonstrations} on a humanoid robot and multiple dual-arm systems. Lastly, we open-source the design of the hardware promoting accessibility and reproducibility. Additional details and open-source information are available at our project website: \url{https://uiuckimlab.github.io/CHILD-pages}.

\end{abstract}
%%%%%%%%%%%%%%%%%%%%%%%%%%%%%%%%%%%%%%%%%%%%%%%%%%%%%%%%%%%%%%%%%%%%%%%%%%%%%%%%

\section{INTRODUCTION}
Teleoperation is a commonly used technique to bridge the gap between robots' current autonomous and physical capabilities. More recently, teleoperation has become a popular method to collect demonstration data for learning-based policies. It is particularly effective when applied to single or dual-arm tasks, allowing robots to perform complex manipulation tasks. However, teleoperating humanoid robots poses unique challenges due to the high degrees of freedom (DoFs), inherent dynamic instability, and the unstructured environments in which they are expected to operate. 

\begin{figure}
    \centering
    \includegraphics[width=1\linewidth]{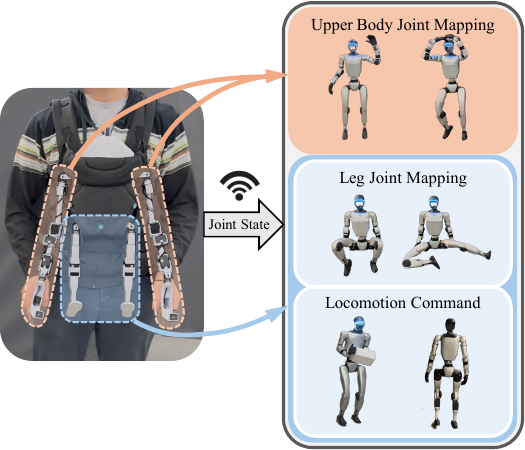}
    \caption{Overview of CHILD humanoid teleoperation \rev{system. Upper body joints are directly mapped, while the legs are directly mapped or used as joy sticks for locomotion control.}}
    \label{fig:first_fig}
\end{figure}

\begin{figure}[!t]
    \centering
    \includegraphics[width=0.85\linewidth]{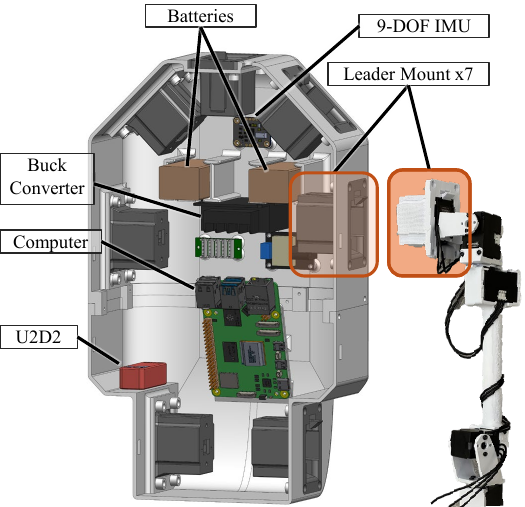}
    \caption{Internal components \rev{and mount locations} of the torso structure.}
    \label{fig:component}
\end{figure}

Joint-level teleoperation using a scaled kinematically identical structure offers many advantages in teleoperation systems, simplifying the controls, providing a direct feedback to the operator on joint limits and singularities, and enabling bilateral operation in the form of force feedback. However, utilizing this method for whole body teleoperation has not been explored extensively. To this end, we introduce the Controller for Humanoid Imitation and Live Demonstration (CHILD), a compact reconfigurable humanoid teleoperation system that offers direct joint-level control over humanoid robots. 
To the best of our knowledge, our proposed device is the first system utilizing the direct joint mapping approach for whole-body control. As illustrated in Fig. \ref{fig:first_fig}, CHILD is a compact system designed to fit in a baby carrier that enables the \rev{user} to teleoperate a wide variety of motions. It is comprised of a small enclosure with seven mounts (Fig. \ref{fig:component}) in which the leader limbs can be easily attached and detached according to the configuration of the follower robot. This reconfigurability enables teleoperation of a variety of robot systems, including full humanoid systems, dual-arm platforms, and single-arm manipulators. By enabling joint-level control and being adaptable to a variety of follower robots, it combines the intuitive advantages of joint-space teleoperation with a general-purpose architecture.
We then validate this system on the \rev{Unitree} G1 humanoid robot \cite{Unitree_g1_2024}, Orthrus \cite{orthrus} a custom dual-arm system mounted on Boston Dynamics' Spot, and a custom kitchen dual arm system \cite{PAPRAS}. 
The key contributions of CHILD are:
\begin{itemize}
    \item \rev{A} whole-body teleoperation system that enables direct, joint-level control of a full-body humanoid or multi-limbed robotic system.
    \item \rev{A} reconfigurable teleoperation system capable of adapting to and controlling a wide range of robot configurations.
    \item \rev{O}pen-source release of the hardware design, promoting accessibility, reproducibility, and further development by the broader robotics community.
\end{itemize}

\section{Related Work}
Many existing works on humanoid teleoperation focus on performing loco-manipulation tasks. In these methods, the upper body is controlled to closely follow the operator's movements, while high-level commands are sent to a lower body controller. The target upper body pose is often estimated using exoskeletons\cite{Homie, ramos_paper} or through vision based systems such as virtual reality (VR)\cite{vr_no_ik, trill, ace, television} \rev{and} RGB-D cameras\cite{anyteleop, dexpilot}. When using vision-based systems, the operator's estimated end effector pose is typically mapped to the robot using inverse kinematics (IK). Such systems enable efficient control of loco-manipulation tasks, but rely on a separate locomotion controller for the lower body, limiting their applications to tasks using the lower body as a mobile base. Because this approach relies only on the end-effector pose, it can be generalized across a wide range of robots and implemented at relatively low cost when using simple RGB or RGB-D cameras. However, this method can be limited in the accuracy of the pose estimation, particularly due to frequent occlusions.
Moreover, the vision based approaches typically provide only unilateral teleoperation—offering no physical feedback to the user about the robot’s interaction with the environment. 
% mention exoskeletons here

% \begin{figure}[!t]
%     \centering
%     \includegraphics[width=1\linewidth]{Figures/control_fig_v2.pdf}
%     \caption{Enter Caption}
%     \label{fig:enter-label}
% \end{figure}

Full-body teleoperation has been accomplished using a variety of methods. He et al.~\cite{omnih20} propose whole body control using a trained control policy with input from a VR headset, voice commands, or an RGB camera. Ishiguro et al.~\cite{tablis} propose the use of a full-body exoskeleton in a seated position to directly control all four limbs. Another approach is to use inertial measurement units (IMUs) placed on the operator \cite{robust-imu, syncronized-imu, imu_retargeting}. When augmented by an additional stability or walking controller, these methods have successfully demonstrated whole-body retargeting and walking. These systems are intuitive but expensive and require the operator to perform all movements, making them unsuitable for long-duration use. Real-time motion capture-based systems\cite{twist, mocap} provide similar capabilities with greater fidelity in pose estimation but are often quite expensive and have similar drawbacks as the IMU based approach. Additionally, motion capture systems are not easily portable, so the teleoperation workspace is limited.

Recently joint-level teleoperation has become a popular approach for dual-arm systems that offers a more direct and interpretable form of control. In recent works such as \cite{Aloha, mobile_aloha, echo, Gello}, operators control scaled models of the follower robots made from inexpensive servo motors. This method simplifies the control pipeline by allowing direct joint state transfer from the leader to the follower. Because they operate at the joint level, they can also support bilateral teleoperation, enabling follower state feedback through mechanisms such as haptics and force feedback. Bilateral teleoperation tends to provide a more intuitive and responsive user experience, particularly in avoiding joint limits or singularities. 

 % \textcolor{red}{add more on mocap?}
% To the best of our knowledge, our proposed device is the first system utilizing the direct joint mapping approach for whole-body humanoid teleoperation. To this end, we introduce Controller for Humanoid Imitation and Live Demonstration (CHILD), a low-cost reconfigurable humanoid teleoperation system that offers direct joint-level control over humanoid robots. As illustrated in Fig. \ref{fig:first_fig}, CHILD is a compact system designed to fit in a baby carrier that enables the operator to teleoperate a wide variety of motions. It is comprised of a small enclosure with seven mounts in which the leader limbs can be easily attached and detached according to the configuration of the follower robot. This reconfigurability enables teleoperation of a variety of robot systems, including full humanoid systems, dual-arm platforms, and single-arm manipulators. By enabling joint-level control and being adaptable to a variety of follower robots, it combines the intuitive advantages of joint-space teleoperation with a general-purpose architecture.
% We then validate this system on the \rev{Unitree} G1 humanoid robot \cite{\rev{Unitree}_g1_2024} and a custom dual-arm system mounted on Boston Dynamics' Spot. 

% However, these methods are usually limited to either stationary or mobile dual-arm systems.

%%%%%%%%%%%%%%%%%%%%%%%%%%%%%%%%%%%%%%%%%%%%%%%%%%%%%%%%%%%%%%%%%%%%%%%%%%%%%%%
%%%%%%%%%%%%%%%%%%%%%%%%%%%%%%%%%%%%%%%%%%%%%%%%%%%%%%%%%%%%%%%%%%%%%%%%%%%%%%%%
%%%%%%%%%%%%%%%%%%%%%%%%%%%%%%%%%%%%%%%%
%%%%%%%%%%%%%%%%%%%%%%%%%%%%%%%%%%%%%%%%
\section{\rev{Hardware Design}}

% \subsection{Hardware Design}
The primary focus for the hardware design is to make a reconfigurable, compact teleoperation system to facilitate whole-body control. The design goals are summarized as follows:
\begin{itemize}
    \item Reconfigurable: The system should be easily customized for different follower robot configurations. We achieve this by implementing seven mounts and a simple clip to attach and detach leader limbs. Off-the-shelf servo motors and 3D printed parts to make the parts relatively low cost and customizable.
    \item Compact: All power, electronics, and teleoperation components are fully contained in a standard baby carrier. This self-contained design enables a single operator to control all four leaders while remaining mobile and able to adjust viewing positions as needed.
\end{itemize}

\begin{figure}
    \centering
    \includegraphics[width=.95\linewidth]{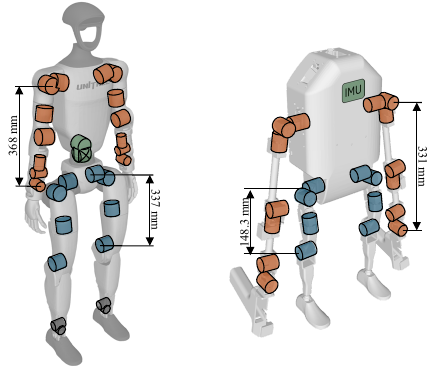}
    \caption{Matching joint configuration and scaling between G1 and CHILD. Best viewed in color.}
    \label{fig:configuration_matching}
\end{figure}
To support teleoperation across a diverse range of configurations, the varying shoulder configurations that are popular among humanoid robots must be considered. The rotation axis of the first shoulder joint is generally inclined between about 0 and 45 degrees. Examples of low inclination shoulder joints include G1\footnote{https://www.Unitree.com/g1} and Boston Dynamics' Atlas\footnote{https://bostondynamics.com/atlas/}, while Figure AI's Figure 02\footnote{https://www.figure.ai/} and Fourier GR-1\footnote{https://www.fftai.com/products-gr1} exhibit inclined shoulder joints closer to 45 degrees. Therefore, CHILD features arm mounts both parallel to the ground and inclined by 45 degrees. In this way, the closest matching set of mounts can be used to approximate the shoulder orientation of the follower. Specifically, it includes a total of seven mounts: two for legs, two for arms parallel to the ground, two for arms offset by 45 degrees, and one on the top to enable neck control.  

The leaders are designed to maintain the same kinematic configuration as the corresponding follower. The translational transformation between each joint coordinate frame is then scaled by a factor $\alpha$ to keep the overall leader workspace well within the operators reach. Consequently, the value of $\alpha$ is adjusted based on the follower robot. For example, we use $\alpha=0.65$ for our custom robotic arm, and $\alpha=0.9$ for the G1 leader arms. Fig. \ref{fig:configuration_matching} displays an example of the joint configuration matching between G1 and CHILD. In this case, the IMU is mapped to the torso joints. A handle with a single-DoF trigger is attached to the end effectors to allow the user to comfortably control parallel grippers. Each joint of the leader limbs utilize relatively low cost servo motors, DYNAMIXEL XL330-M288-T. Along with providing high-resolution encoders for joint state measurements, they also enable force feedback. 

The outer dimensions of the torso and the mount locations are primarily limited by the size of the baby carrier and the gaps between the straps. The width of the torso is selected to be 90\% of the shoulder width of the humanoid robot, with the two bottom mounts inset to maintain the same shoulder to hip ratio.  
To achieve reconfigurability of the system, each mount features a print-in-place compliant retaining clip to hold each leaders’ base servo motor securely in place. Additionally, pogo pin connectors are used to transmit power and communication from the torso to the leaders, eliminating the need to manage wiring. This allows the leaders to be quickly attached or detached, making it suitable for different robot followers. 
% Additionally, this allows the leader limbs to be attached after placing the shell in the baby carrier 
 
\begin{figure}[!t]
    \centering
    \includegraphics[width=1\linewidth]{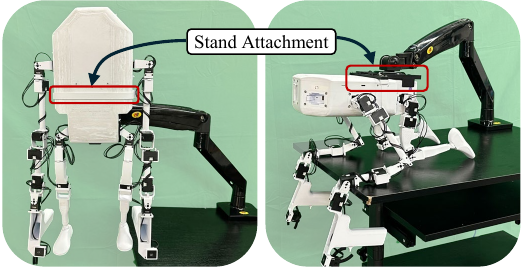}
    \caption{CHILD mounted on a six-DoF monitor stand.}
    \label{fig:stand-setup}
\end{figure}

Utilizing the baby carrier allows for a single operator to directly control up to two limbs at a time in an approximately upright orientation. In cases where the task requires more complex motions or would require the operator to be in an uncomfortable orientation such as for a crawling motion, CHILD can also be deployed on a stationary platform off of the body. For such cases, we design a desktop monitor stand adapter as shown in Fig. \ref{fig:stand-setup}. Paired with a \rev{standard} six-DoF monitor arm stand, the CHILD can be oriented in any pose desired, and multiple operators can simultaneously control all four limbs.
\rev{The entire system is} 3D printed using PLA on an FDM printer, making the hardware relatively low cost and easily reproducible. \rev{The total cost of the system amounts to just under \$1k, with the full bill of materials available on the project github}.

% \begin{figure}
%     \centering
%     \includegraphics[width=1\linewidth]{Fig1_collage.png}
%     \caption{Enter Caption}
%     \label{fig:enter-label}
% \end{figure}
% \begin{figure}[!t]
%     \centering
%     \includegraphics[width=1.0\columnwidth]{Figures/baby_dimensions_fig.pdf}
%     \caption{Key dimensions. Also showing the rotation axes for leader mounts}
%     \label{fig:control-diagram}
% \end{figure}
\subsection{Electronics}
Mobility is essential for using this system in a diverse set of tasks. Therefore, we design it to contain all power and electronics within the torso shell, and communicate with the follower robot via Wi-Fi. The onboard computer and motors are powered by two 4S 75C LiPo batteries stepped down using a 5V 15A buck converter. Cumulatively, this provides a runtime of continous teleoperation of over two hours. A Raspberry Pi 5 is used as the onboard computer to read and send the leader joint positions and velocities. Communication between the onboard computer and the motors is established via the DYNAMIXEL U2D2. Finally, torso orientation is measured using a BNO055 9-axis IMU, also located in the torso, and communicates to the onboard computer using the I2C protocol. The IMU contains an onboard sensor fusion algorithm to provide the absolute orientation from the raw accelerometer, gyroscope, and magnetometer readings. The internal layout of the electronics is shown in Fig. \ref{fig:component}. 

% \begin{table}[!h]
% \centering
% \caption{\rev{CHILD approximate cost breakdown}}
% \label{tab:bom}
% \begin{tabular}{|c|c|}
% \hline
% \textbf{Items}              & \textbf{Cost (\$)}              \\ \hline
% Dynamixel Motors                 & 630                \\ \hline
% Electronics (Computer, IMU, etc.)           & 266            \\ \hline
% Other Hardware (Baby Carrier, 3D prints, etc.)      & 70         \\ \hline
% \end{tabular}
% \end{table}

% \begin{table}[t]
% \caption{Electrical Component List}
% \label{tab:electrical components}
% \centering
% \begin{tabular}{ccccc}
% \toprule
% \textbf{Direction} & \textbf{Sign} & \textbf{Recovered} & \textbf{Non-recoverable} & \textbf{No Fall} \\
% \midrule
% \multirow{2}{*}{x Perturbation} & $+$ & 7.35 & 11.27 & 2.45 \\
%                                 & $-$ & 13.73 & 0.98 & 2.94 \\
% \multirow{2}{*}{y Perturbation} & $+$ & 7.35 & 2.94 & 4.90 \\
%                                 & $-$ & 8.82 & 2.94 & 1.96 \\
% \multirow{2}{*}{z Perturbation} & $+$ & 6.37 & 7.84 & 0.98 \\
%                                 & $-$ & 5.88 & 10.78 & 0.49 \\
% \midrule
% \textbf{Total} & & \textbf{49.51} & \textbf{36.76} & \textbf{13.73} \\
% \bottomrule
% \end{tabular}
% \end{table}

\begin{figure*}[!t]
    \centering
    \begin{subfigure}{\textwidth}
        \centering
        \includegraphics[width=1\linewidth]{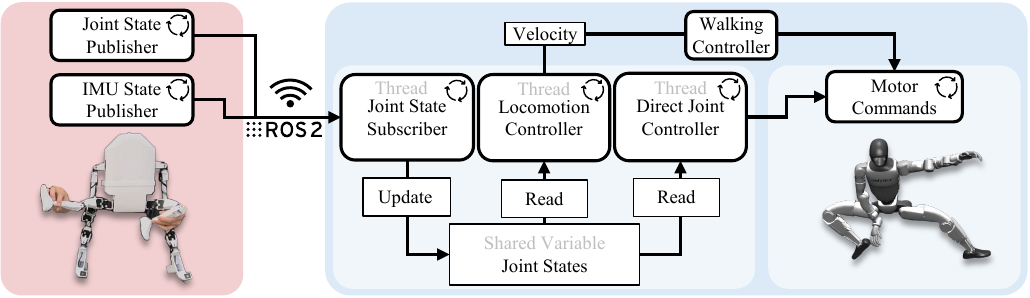}
    \end{subfigure}
    \caption{Overview of the teleoperation software architecture.}
    \label{fig:software_diagram}
\end{figure*}

\begin{figure}[!t]
    \centering
    \includegraphics[width=1\linewidth]{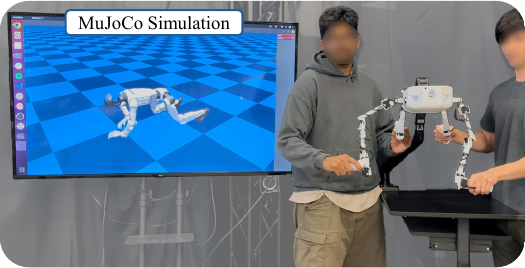}
    \caption{Simulation of a crawling motion using CHILD \rev{and two operators}.}
    \label{fig:crawl_motion}
\end{figure}

\section{\rev{Software}}

\subsection{Control Interface}
The control interface also adopts a reconfigurable design. To teleoperate the humanoid robot using the attached leaders, users only need to write configuration files that specify the motor-joint mapping on the leader, and which joints on the leader will be mapped to the follower robot.
The overall control architecture is illustrated in Fig. \ref{fig:software_diagram}. 
Based on the given configuration files, the computer inside the torso shell reads joint states and IMU data from the leader devices and publishes the information over Wi-Fi using ROS2.
A separate onboard computer in the humanoid robot subscribes to the leader states and sends motor commands to the corresponding actuators.

Inside of the onboard computer, three modules run asynchronously, as illustrated in Fig. \ref{fig:software_diagram}. The first module is \texttt{Joint State Subscriber}, which subscribes to the joint states and IMU data from CHILD, and update the \texttt{Joint States} variable. 
This variable is shared with other two modules, \texttt{Locomotion Controller} and \texttt{Direct Joint Controller}.
These controllers read the \texttt{Joint States} variable, and translate it into motor commands.
All modules run asynchronously on separate threads, sharing the variable of joint states to ensure non-blocking operation and support real-time, low-latency control, \rev{achieving an average latency of about 14ms}.

The \texttt{Direct Joint Controller} maps the leader joint states directly onto the follower joint states, and converts IMU data to euler angles, which can then be directly sent as position commands to the three torso joints. 
This controller is mainly used for upper body (arms and torso) control.
% This direct joint-to-joint mapping enables intuitive teleoperation. 
We can also use this controller for the lower body (legs) to achieve a full-body teleoperation system. In this mode, the humanoid legs directly follow the joint positions of the leader legs. This enables full-body teleoperation, allowing the humanoid to perform a wide range of motions and poses.

\texttt{Locomotion Controller} supports a different type of control for lower body.
This module translates leader joint states into velocity messages and sends them to a walking controller that manages balance and locomotion.
This flexible control system allows the humanoid to perform a wide variety of tasks beyond simple motion replication—for example, pick-and-place operations across the environment. 
For our experiment with the humanoid robot, we used the embedded walking controller, but it can be replaced with a custom controller if needed.

Utilizing the modular control architecture, a simulation can easily be integrated in place of the follower robot hardware. A separate node running a MuJoCo simulation subscribes to the leader joint states and sends to joint commands to the simulation. A simulation environment provides more flexible control over the use of the leader IMU. We use the IMU to define the orientation of the follower robot's torso relative to the world frame, rather than as waist joint commands. With this method, we can orient the robot to perform motions such as crawling.

% Since the leaders are designed as scaled versions of the followers, the teleoperation control can be made to be quite simple. In contrast to vision based or other such teleoperation methods, the joint position and velocities can be directly mapped to the follower robot, which enables the operator to understand limitations of the robot such as singularities more easily. This also eliminates the need for forward and inverse kinematics, which can introduce errors in the joint positions, particularly for 7 DoF manipulators. To decrease vibrations on the follower side, the joint commands are first filtered using a low pass filter. This contributes a delay of \textcolor{red}{\#}ms between the operator’s input and the robot motion. The orientation of the torso is controlled by the IMU. The absolution orientation quaternion is converted to Euler angles, which can then be directly sent as position commands to the three torso joints. 

% \begin{figure}[!t]
%     \centering
%     \includegraphics[width=1.0\columnwidth]{control_fig.png}
%     \caption{Control diagram for teleoperation.}
%     \label{fig:control-diagram}
% \end{figure}
\subsection{Teleoperation Session}
The overall teleoperation procedure follows the structure described in \cite{paprle}. 
After launching the teleoperation system, the controller waits for the operator to activate the session by holding the grippers on the leader device for three seconds. 
Once triggered, the robot slowly moves to the current pose of the leader device to synchronize their positions. After synchronization, the robot continuously follows commands from the leader device. 
For additional details on the teleoperation framework, please refer to \cite{paprle}.

In the case of CHILD, we introduce an additional mode-switching mechanism within the teleoperation session. 
When the lower body of the humanoid is controlled in locomotion mode, the user can activate joystick-based control by closing a single gripper (either left or right) for one second. 
\rev{The corresponding arm is then deactivated while the leg functions as a joystick, with its hip joint angles interpreted as velocity commands for the walking controller; roll, pitch, and yaw mapping to left/right, forward/backward, and rotational velocities respectively.} 
If this mode is not activated, the system defaults to outputting zero velocity. \rev{The operator can reactivate the arm by closing the gripper for one second.} 

\begin{figure*}[!t]
    \centering
    \begin{subfigure}{\textwidth}
        \centering
        \includegraphics[width=\textwidth]{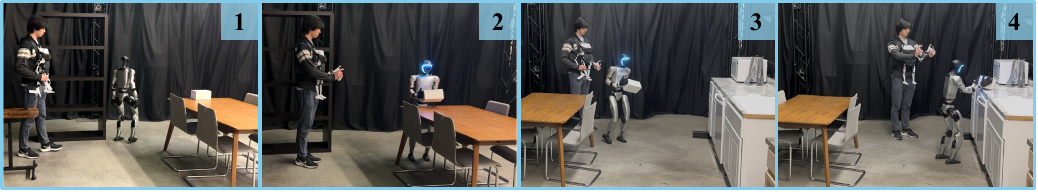}
        \caption{Loco-manipulation mode to pick and place a box}
        \label{fig:box_carry}
    \end{subfigure}
    % \vspace{2mm}
    \begin{subfigure}{\textwidth}
        \centering
        \includegraphics[width=\textwidth]{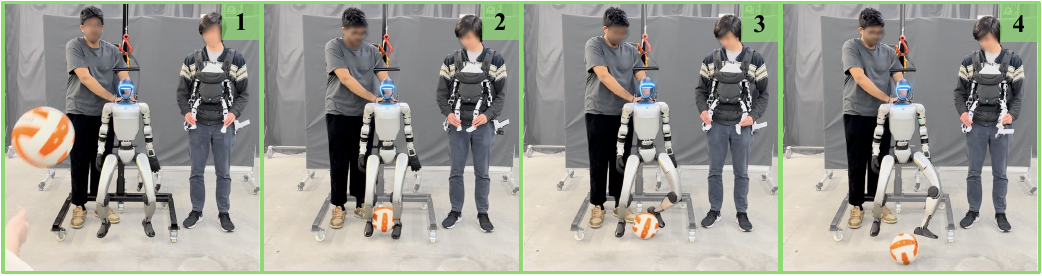}
        \caption{Direct joint teleoperation of legs to catch and pass a ball}
        \label{fig:ball_catch}
    \end{subfigure}
    \caption{Example tasks using two different modes of humanoid teleoperation}
    \label{fig:g1_tasks}
\end{figure*}
    
\begin{figure}[!t]
  \begin{subfigure}[b]{0.43\columnwidth}
    \centering
    \includegraphics[width=\linewidth]{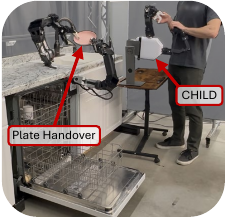}
    \caption{Kitchen dual-arm system}
    \label{fig:kitchen}
  \end{subfigure}
  \begin{subfigure}[b]{0.57\columnwidth}
    \centering
    \includegraphics[width=\linewidth]{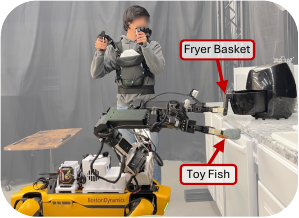}
    \caption{Mobile dual-arm system}
    \label{fig:orthrus}
  \end{subfigure}
  \caption{Examples using CHILD for non-humanoid robot systems
    % \subref{fig:graph} The factor graph illustration.
    % \subref{fig:factor} The constraint points definition for the object pose factor.
  }
  \label{fig:alternate_configs}
\end{figure}
\subsection{Adaptive Force Feedback}
To improve the operator experience and avoid singular joint configurations, force feedback terms are applied through commanded joint torques. Since the operator can only directly control the six-DoF of the end effector, they cannot reliably control all seven DoFs of the leader arms. For this reason, a virtual spring is applied to each joint to bias the joint positions towards a specified base position. \rev{The bias torque is calculated as
\begin{equation}
% \mathbf{\tau}_\text{bias}
\boldsymbol{\tau}_{\text{bias}} = \mathbf{k}\left(\mathbf{q}(t) - \mathbf{q}^{\text{base}}\right),
\end{equation}
where  $\mathbf{q}^{\text{base}}$ is the base position, $\mathbf{q}(t)$ is the current joint positions, and $\mathbf{k}$ is a diagonal matrix containing the user-defined spring constant for each joint. }
The strength of the force feedback varies depending on the phase of the teleoperation session. For example, when the operator is using a leg as a joystick to navigate the humanoid, stronger force feedback is applied to the deactivated arm. \rev{This helps the operator return the leader arm to its previous position before switching control modes back from locomotion control.}

% For many loco-manipulation tasks, direct joint level control of the lower body is unnecessary, and even complicates the task unnecessarily. Many pick and place, assembly, or manufacturing tasks require manipulation using the arms, while the lower body is simply used as a mobile base. To demonstrate the capability of this system in these scenarios, we take an approach similar to HOMIE\cite{Homie}; a dedicated walking controller is used for the lower body while the upper body is controlled at the joint level via the leader arms. 

% For more complex movements, it may not be sufficient to use a dedicated controller for the lower body. In this case, the CHILD platform allows the operator to control the full joint state of the robot. 

% \begin{figure}[!t]
%     \centering
%     \includegraphics[width=1\linewidth]{control_fig_temp.png}
%     \caption{Enter Caption}
%     \label{fig:enter-label}
% \end{figure}

%%%%%%%%%%%%%%%%%%%%%%%%%%%%%%%%%%%%%%%%%%%%%%%%%%%%%%%%%%%%%%%%%%%%%%%%%%%%%%%%
\section{Results}
%%%%%%%%%%%%%%%%%%%%%%%%%%%%%%%%%%%%%%%%%%%%%%%%%%%%%%%%%%%%%%%%%%%%%%%%%%%%%%%%
In this section, we provide several demonstrations using multiple follower platforms to validate the system. 
We tested our system on the following robots:      
\begin{itemize}
    \item \rev{G1}
    \item Orthrus (a dual-arm mobile manipulator) \cite{orthrus}
    % \item Mobile Manipulator with Sensor Module Head
    \item Dual-arm kitchen setup \cite{PAPRAS}
\end{itemize} 
The accompanying video includes the full demonstrations described in this section.
\subsection{Humanoid Control}
\subsubsection{Upper body teleoperation} In this experiment, we demonstrate loco-manipulation capabilities of this system by using the direct joint controller for the upper body and a locomotion controller for the legs. We utilize the on-board walking controller for lower body walking and stability control. Shown in Fig. \textcolor{red}{\ref{fig:box_carry}}, we use this control framework to navigate between a table and book shelf (1), grasp a box off of the table (2), transport it to a second table (3), and \rev{set it down} (4). 

\subsubsection{Full-body teleoperation}
To demonstrate full-body joint-level teleoperation, we perform several movements requiring fine control over all of the robot's joint states. Regarding the leg leaders, we reduce the degrees of freedom to make the system easier to control. Specifically, we choose to neglect the ankle joints. In Fig. \ref{fig:ball_catch}, we take advantage of the low latency and fine control over the leg joint states to catch a ball with the feet (2), drop the ball to the ground (3), and pass the ball back to the thrower (4). Note: the humanoid robot is being supported by a gantry and a person. Next, in Fig. \ref{fig:crawl_motion}, we demonstrate a crawling motion in simulation. The pelvis of the robot is fixed in place, with the orientation controlled by the leader IMU. The waist joints are consequently fixed. Due to the increased complexity of controlling four limbs, we perform this demonstration with the device mounted on a monitor stand and two operators.
% \subsection{Dual-arm Mobile Manipulation}

\subsection{Alternate Configurations}
    In addition to humanoid robots, the reconfigurability of the proposed system allows for the teleoperation of other forms of robotic arm systems. For example, by utilizing the top mount and one side mount, we can approximate the configuration of a custom dual-arm kitchen setup as shown in Fig. \ref{fig:kitchen}. Using the same control framework, we can perform manipulation tasks in the kitchen environment, including loading the dishwasher. In Fig. \ref{fig:kitchen} we show an intermediate step of this task, handing over a plate. 
    We also test this system on Orthrus. In this case, the two arms are mounted at an inclination of 45 degrees, so we utilize the corresponding mounts on the leader system. In Fig. \ref{fig:orthrus}, we show an intermediate step of the following pick and place task: grasp the toy fish, open the air fryer basket, place the fish inside, and close the air fryer. Note: in this demonstration, the mobile base, Spot, is operated through a handheld controller off-screen and is not integrated with the teleoperation framework.  
    
% \subsection{Something for neck}

%%%%%%%%%%%%%%%%%%%%%%%%%%%%%%%%%%%%%%%%%%%%%%%%%%%%%%%%%%%%%%%%%%%%%%%%%%%%%%%%
\section{Conclusions and \rev{Discussions}}
% A reConfigurable Humanoid Interface for whole-body control
In this paper, we introduce a humanoid teleoperation device CHILD, Controller for Humanoid Imitation and Live Demonstration, which leverages direct joint state mapping to achieve full-body control. The device features several leader mounts to target the control of many different humanoid configurations. We apply the proposed device to demonstrate full-body teleoperation of a humanoid and dual-arm control on a mobile manipulator and custom kitchen environment. We demonstrate the utility of the reconfigurable design of CHILD to control a custom kitchen dual-arm system. Moreover, we show the capability of this device to perform complex motions, taking advantage of direct joint level teleoperation for full-body control.

% \rev{
%  During full body joint level teleoperation, the follower robot must be externally supported, limiting the possible applications using this mode. Additionally, dexterous hands are not yet supported, limiting the range of manipulation tasks that can be accomplished using this device.}
\rev{In future works, we intend to address several limitations in the hardware and controls.}
\rev{Regarding the system hardware,} the current leader end effectors are simple triggers that are best used for parallel grippers. Higher DoF grippers can be designed to control dexterous hands and enable more complex loco-manipulation tasks. \rev{From a control standpoint, the humanoid robot requires external support for whole body joint level control, limiting the diversity of tasks that can be performed in this mode.} Stability controllers can be developed to allow full joint-level teleoperation while the robot is free-standing. Additionally, such controllers can improve the stability during loco-manipulation tasks.
% stronger motors for the shoulder and elbow joints on the leader arms to allow for stronger force feedback.
% place while the operator is controlling the legs. 

% \rev{Currently, the walking controller does not account for changes in the arm pose or mass resulting from holding a payload, increasing the difficulty in locomotion control for the operator.}

%%%%%%%%%%%%%%%%%%%%%%%%%%%%%%%%%%%%%%%%%%%%%%%%%%%%%%%%%%%%%%%%%%%%%%%%%%%%%%%%
\bibliographystyle{IEEEtran}
\bibliography{references.bib}
\newpage
\end{document}